\documentclass[sigconf]{acmart}

\usepackage{algorithm2e}
\usepackage{grffile} 
\usepackage{xcolor}
\usepackage{supertabular} 
\usepackage{cleveref}
\usepackage{multirow}
\usepackage{listings}
\usepackage{setspace}

\usepackage{newfloat}
\DeclareFloatingEnvironment[
   fileext=lol,
   name=Listing
]{listing}

\usepackage{natbib}

\usepackage{subcaption}
\DeclareCaptionSubType{listing}

\usepackage{mathtools} 
\newtheoremstyle{sltheorem}
{}                
{0pt}             
{\slshape}        
{}                
{\bfseries}       
{.}               
{ }               
{}                

\theoremstyle{sltheorem}

\newtheorem{challenge}{Challenge}
\AfterEndEnvironment{challenge}{\noindent\ignorespaces}

\newtheorem{capability}{Capability}
\AfterEndEnvironment{capability}{\noindent\ignorespaces}

\newtheorem{Opportunity}{Opportunity}
\AfterEndEnvironment{Opportunity}{\noindent\ignorespaces}

\copyrightyear{2020}
\acmYear{2020}
\setcopyright{acmcopyright}\acmConference[ICAIF '20]{ACM International Conference on AI in Finance}{October 15--16, 2020}{New York, NY, USA}
\acmBooktitle{ACM International Conference on AI in Finance (ICAIF '20), October 15--16, 2020, New York, NY, USA}
\acmPrice{15.00}
\acmDOI{10.1145/3383455.3422564}
\acmISBN{978-1-4503-7584-9/20/10}

\acmConference[ICAIF '20]{ICAIF '20:
First ACM International Conference on AI in Finance}
{October 15--16, 2020}{New York}

\usepackage{enumitem}
\usepackage{lipsum}
\setlist[itemize]{leftmargin=*}

\begin{document}

\title{Towards Self-Regulating AI: Challenges and Opportunities of AI Model Governance in Financial Services}

\author{Eren Kurshan}
\affiliation{
  \institution{Columbia University}
  \city{New York}
  \state{New York}
}
\email{ek2529@columbia.edu}

\author{Hongda Shen}
\affiliation{
  \institution{University of Alabama}
  \city{Huntsville}
  \state{Alabama}
}
\email{hs0017@alumni.uah.edu}

\author{Jiahao Chen}
\orcid{0000-0002-4357-6574}
\affiliation{
  \institution{J.\ P.\ Morgan AI Research}
  \city{New York}
  \state{New York}
}
\email{jiahao.chen@jpmorgan.com}


\begin{abstract}
AI systems have found a wide range of application areas in financial services. Their involvement in broader and increasingly critical decisions has escalated the need for compliance and effective model governance. Current governance practices have evolved from more traditional financial applications and modeling frameworks. They often struggle with the fundamental differences in AI characteristics such as uncertainty in the assumptions, and the lack of explicit programming. AI model governance frequently involves complex review flows and relies heavily on manual steps. As a result, it faces serious challenges in effectiveness, cost, complexity, and speed. Furthermore, the unprecedented rate of growth in the AI model complexity raises questions on the sustainability of the current practices. This paper focuses on the challenges of AI model governance in the financial services industry.  As a part of the outlook, we present a system-level framework towards increased self-regulation for robustness and compliance. This approach aims to enable potential solution opportunities through increased automation and the integration of monitoring, management, and mitigation capabilities. The proposed framework also provides model governance and risk management improved capabilities to manage model risk during deployment.
\end{abstract}

\keywords{Financial services, model governance, model risk management, machine learning, artificial intelligence}

\settopmatter{printfolios=true} 
\maketitle


\section{Introduction}
\label{sec:intro}



In recent years, AI adoption in the financial services industry has grown significantly. AI applications span a wide range of business functions such as new product development, business operations, customer service and client acquisition \citep{Heaton16,Neurips19}. The range and criticality of the decisions made by AI have increased the need for compliance and effective model governance.

\begin{figure}
\centering
\vspace{5pt} 
\includegraphics[keepaspectratio,width=0.75\columnwidth]{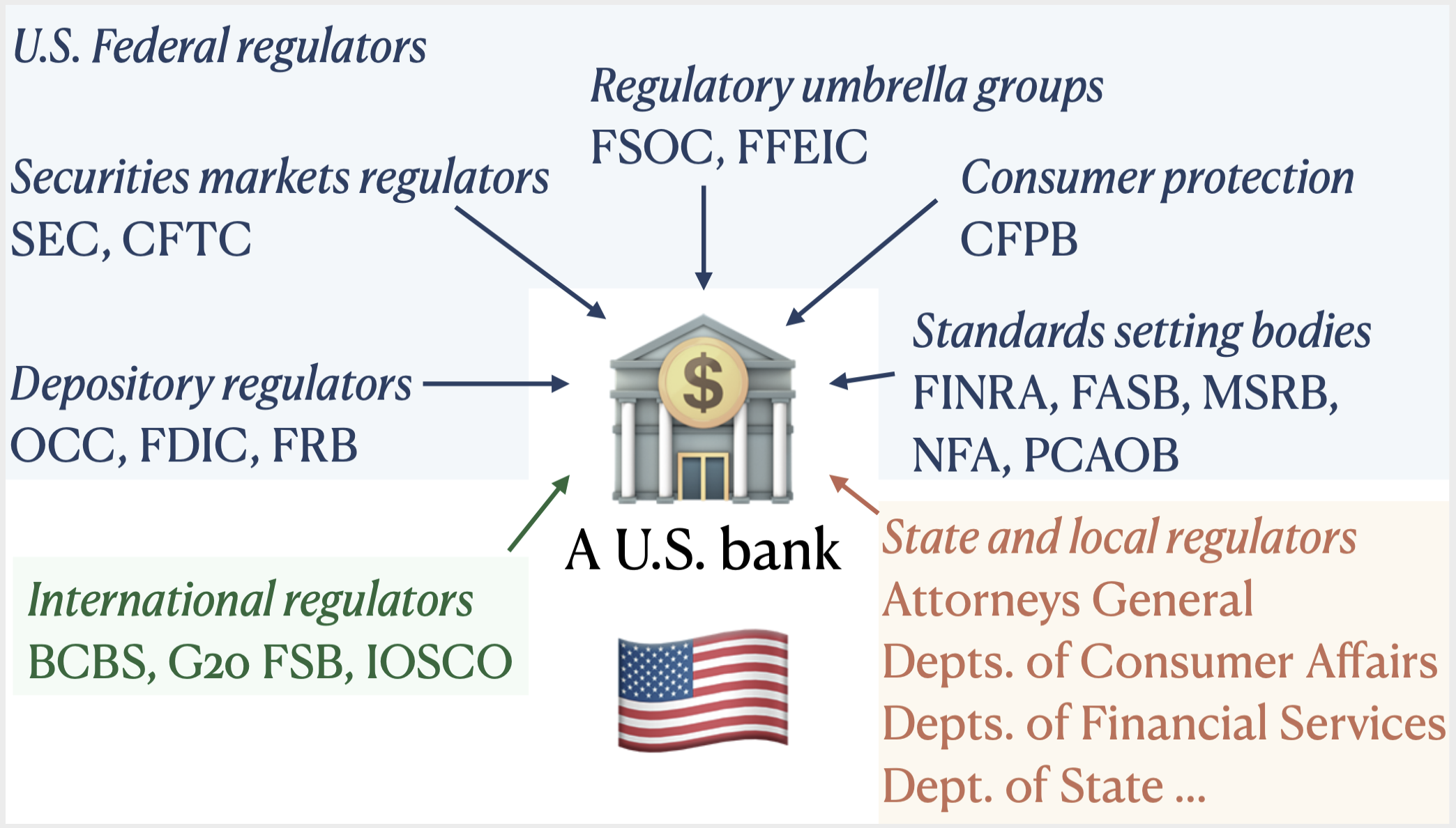}
\setlength{\belowcaptionskip}{-14pt}
\caption{\small{Financial institutions in the US\ are regulated by a number of regulatory entities at local \textit{\small {(yellow)}}, state \textit{\small {(yellow)}}, federal \textit{\small{(blue)}} and international levels} \textit{\small{(green)}} \citep{R44918}.}
\label{fig:regulators}
\end{figure}

In financial services, all models are required to go through an internal risk management and regulatory compliance process \citep{Harvard16}. As shown in \Cref{fig:regulators}, this was historically built to address the requirements of a complex system of regulatory entities. 
Since its origins from the compliance with regulations from the Office of the Comptroller of the Currency (OCC) and  CRD IV, CRR  in the European Union,
this risk management and compliance process has evolved with modeling frameworks used in finance.
The resulting process struggles with the differences in the underlying characteristics of the AI models such as intrinsic learning from the data without explicit programming and the lack of \textit{a priori} assumptions.  

Traditional financial modeling frameworks produce more transparent models, in which the outputs have clear conditional dependencies on the model parameters and variables. Hence, the model governance reviews on these variables, parameters, and their relationships are more effective.
In contrast, recent modeling frameworks more often produce a \textit{model} as the output, with intrinsically encoded data representations and no clear dependencies between the model parameters and the outputs. As a result, current governance practices struggle with AI models, regardless of how rigorous the reviews are, and face serious issues in terms of effectiveness, agility, cost, and complexity. It is unclear if the fine-grained and manual reviews are sustainable with the current rates of growth in the AI model complexity. Model governance requires more system-level evaluation for the AI models, as well as more data and behavior-centric analysis. 

 In recent months, the AI models in financial services have gone through a natural stress test due to the global pandemic. During this period, most models experienced serious accuracy and reliability issues \cite{mckinsey}. Financial firms struggled with recalibrating and retraining their existing AI models, as well as trying to rapidly build new ones. Many of these efforts faced serious challenges, due to limited automation and the complexities of the development and governance processes \cite{mckinsey}. However, they built up the interest and urgency to develop more robust AI systems and upgrading model governance solutions. 
 
 Overall, these factors motivate a strategic change in AI governance, from the existing sequential manual processes towards \textit{a system-level approach with increased self-regulation capabilities through continuous monitoring and mitigation}. This can be achieved at the boundary of the AI and governance system designs. This paper has 3 primary goals:

\begin{itemize}[topsep=0pt]
\item{Overview some of the common challenges in AI model governance and start an initial conversation on the pain points and solution opportunities.}
\item{Present an AI system framework that enables the integration and automation of key governance functions towards increased self-regulation.}
\item{Discuss the potential solution opportunities as well as adoption considerations.}
\end{itemize}
The paper is structured as follows: \Cref{sec:model-gov} is a high-level overview of model risk management; \Cref{sec:challenges} discusses some of the practical issues in AI governance; \Cref{sec:controller} presents an AI system framework and building blocks; \Cref{sec:opportunities} overviews the potential solution opportunities; and \Cref{sec:Adoption} discusses the adoption considerations.
This paper \textit{does not} aim to propose the next-generation risk management framework for the financial services industry.
Rather, it explores challenges and novel solution approaches specifically for AI models, from a model and system development perspective. Also, it does not aim to fully automate governance processes or achieve complete self-regulation. Instead, it highlights opportunities to improve the AI
system design and streamline the process through increased automation. In this paper, we use a broader definition of model governance which covers the end-to-end process, including but not limited to, model risk management. 

\subsection{Overview of the Model Governance Process}
\label{sec:model-gov}



Model risk management aims to identify and minimize the risks associated with the models used in the financial services industry.
At a high level, it aims to identify key risks, assess, minimize, and monitor them over the model's lifecycle.
The process has many steps covering the development, implementation, testing, and deployment stages.
The exact processes vary significantly across the industry.
Some of the common steps include (but are not limited to):
(1) \textit{Model Risk Rating}: Initial risk assessment guides the entire governance process and typically considers the materiality and the model characteristics. Higher risk models go through more rigorous and extensive reviews.
Model use review ensures that the model is used for the intended use case and matches the requirements.
(2)\textit{ Initial Model Validation}: (2.a)\textit{ Model Assessment:} Reviews the model in terms of the reasonableness of the overall modeling framework. It typically includes an assessment of the application characteristics, assumptions, and data inputs. (2.b) \textit{Process Verification:} High-level variable selection, model code, and data sources are verified. (2.c)\textit{ Results Analysis:} Evaluates the performance of the model. Frequently involves back-testing, benchmarking and other evaluations.
(3)\textit{ On-Going Monitoring}: Supervises the model behavior periodically during the deployment. Periodic reviews of its outputs are planned based on the model risk classification (quarterly, semi-annually, yearly), where the higher risk models are reviewed more frequently. 



\section{AI Model Governance Challenges}
\label{sec:challenges}


\subsection{Governance Process Design}

\begin{challenge}
Model governance was designed for traditional financial models, not AI.
\end{challenge}
As discussed in \Cref{sec:intro}, model governance has its historical roots in traditional financial models such as capital analysis and asset pricing.
The resulting governance processes were designed around statistical models with clear assumptions, well-established relationships among variables, and explicit programs.
More recent modeling frameworks exhibit fundamentally different characteristics such as learning without explicit programming, inherent opacity, and conditional uncertainties.
These dissimilarities limit the effectiveness of the current practices.
Studies have highlighted the need to customize the governance process to focus more on AI-specific criteria and risks \citep{EY_AI,HKIMR}.

\begin{figure}
\centering
\includegraphics[keepaspectratio,width=0.8\columnwidth]{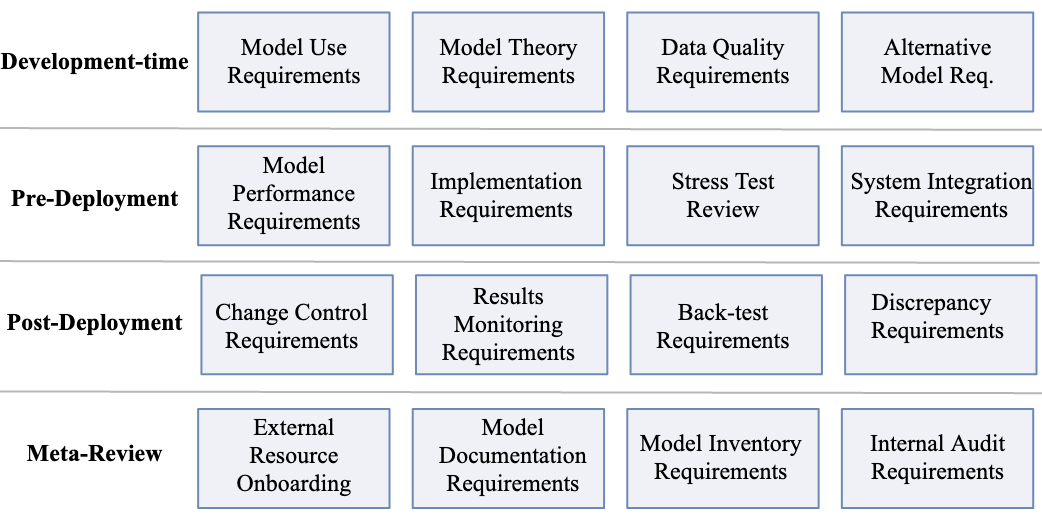}
\setlength{\belowcaptionskip}{-8pt} 
\caption{\small Regulatory requirements by the Office of the Comptroller of the Currency (OCC) over the model's lifecycle.}
\label{fig:OCC-2011}
\end{figure}

\begin{challenge}
Regulatory requirements span the full model lifecycle, while governance feedback mostly focuses on post-development.
\end{challenge}

\Cref{fig:OCC-2011} shows some of the regulatory checks mandated by the OCC, covering the entire model development and deployment timelines.
Nevertheless, model governance reviews typically focus on later, post-development stages,
whose timing is partially driven by the model review backlogs and bandwidth limitations of the model risk management teams. The lack of automated governance tools and feedback during the earlier stages of development causes potential risks and concerns to be identified in later stages. This delays the development timelines and causes restarts to the development and governance processes.

\subsection{Manual Processes and Reviews}

\begin{challenge}
Each AI governance review is unique and has subjective variations.
\end{challenge}
In recent years, financial institutions have taken substantial steps towards standardizing their internal model risk management procedures.
Many firms offer high-level guidelines, process templates and documents.
However, governance procedures still rely on custom steps decided on a case-by-case basis.
Often, the individual governance steps, target metrics, and KPIs are customized for each model.
Governance reviews vary from model to model, governance team to team, and even over time, subject to the available capacity of risk management teams.
As a result, the outcomes of the review process inherit some subjectivity, which makes it difficult to standardize, and is more prone to errors and human oversights \citep{EY_AI}. 

\begin{figure}
\centering
\includegraphics[keepaspectratio,width=0.75\columnwidth]{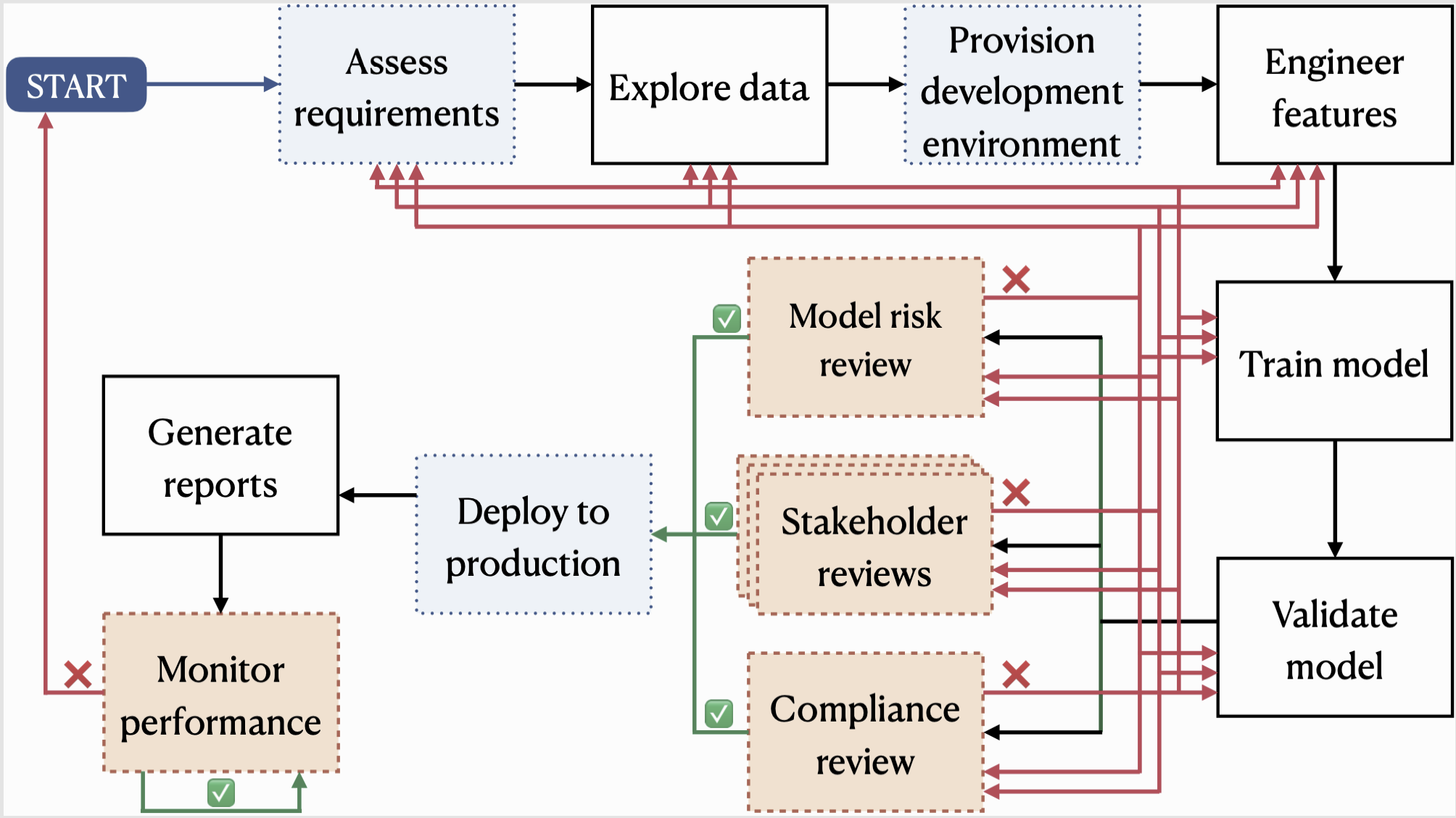}
 \setlength{\belowcaptionskip}{-4pt} 
\caption{\small A simplified model development workflow in financial services, \small{including standard development steps, production steps \textit{\small{(blue)}}, regulatory steps \textit{\small{(yellow)}}, and numerous feedback loops \textit{\small{(green and red)}}}.
}
\label{fig:model-dev-workflow}
\end{figure}

\begin{figure}
\centering
\includegraphics[keepaspectratio,width=0.75\columnwidth]{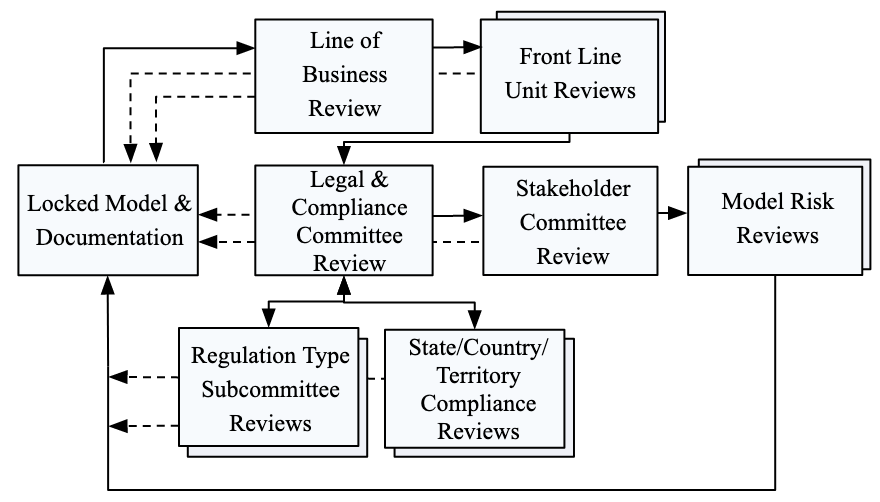}
\setlength{\abovecaptionskip}{2pt} 
\setlength{\belowcaptionskip}{-12pt} 
\caption{\small A sample partial view of AI model governance committee reviews.}
\label{fig:GovernanceCommittees}
\end{figure}

\begin{challenge}
Model governance workflows often have complex and interlacing manual stages.
\end{challenge}
\noindent AI governance reviews frequently follow complex flow diagrams with numerous review stages and interdependencies, representing separation of coverage among aspects of business, risk management, and compliance \citep{Engdahl2014}. The resulting process can take a year or even longer.
\Cref{fig:model-dev-workflow,fig:GovernanceCommittees} illustrate parts of development and model committee review flows respectively. Though the exact process may vary from firm to firm, it frequently involves elaborate committee reviews with many stakeholders represented, like model development, deployment, model risk management, audit, line of business, legal, and compliance.
Each team serves a different role such as model developer, owner, user, stakeholder, administrator, monitoring unit, risk committee, as well as representing their organization's unique business strategies, goals, and requirements.
Each group or committee review may be recursive, incurring further nested reviews. For instance, within the legal and compliance review, regulatory subcommittees such as the Fair Lending Act Committee, country, and territory committees may each perform nested reviews.
While they can make significant changes to the model, each review requires a final and locked version of the model and its documentation.
Any significant change recommended by a committee is performed by the model development team and restarts the process with an updated and locked model.

\begin{challenge}
Defining accountability in the complex governance process flows is difficult.
\end{challenge}

Complex review flows exacerbate the underlying accountability issues in AI models.
The interlacing governance flows inherently distribute the decision making power pervasively among numerous stakeholder organizations.
At times, the strategic goals and risk appetites of the teams involved may conflict.
Business organization strategies may align with higher risk and performance modeling decisions, while risk and compliance organizations may require the opposite.
During the sequential review flows, one committee's recommendations may even conflict that of others. 
As a result, the accountability of the resulting decisions gets locked into the underlying complexities of the process.

\subsection{Process Cost, Fines and Penalties}
\begin{challenge}
Current governance practices translate to high compliance costs. 
\end{challenge}
Compliance processes are usually highly manual functions in the financial services industry. Last year, US financial firms spent around US\$80 billion for compliance purposes, which is expected to grow by 50\% to US\$120 billion in 2025 \citep{ThomsonReuters18}.
Despite the increase in spending, compliance remains a challenge.
Human-based compliance processes and the lack of automation have been considered among the top reasons for the high compliance costs.
As a result, regulatory technology offerings have gained traction in recent years.
However, integrating, scaling and calibrating third-party vendor solutions have been demanding exercises.

\begin{challenge}
Despite being considered the standard and safest path, 
manual governance processes still result in high regulatory penalties.
\end{challenge}
In addition to the high cost of compliance and governance, the resulting process is challenged with high penalties in the form of growing regulatory fines.
US financial institutions paid close to \textit{US\$320 billion} in regulatory fines in the years 2008--2016 \cite{BCG}.
This raises questions on the effectiveness of the current manual governance practices.


\subsection{AI Model Complexity}

\begin{challenge}
AI model complexity is growing exponentially, making fine-grained governance reviews infeasible.
\end{challenge}
\begin{figure}[ht!]
\centering
\includegraphics[keepaspectratio,width=0.75\columnwidth]{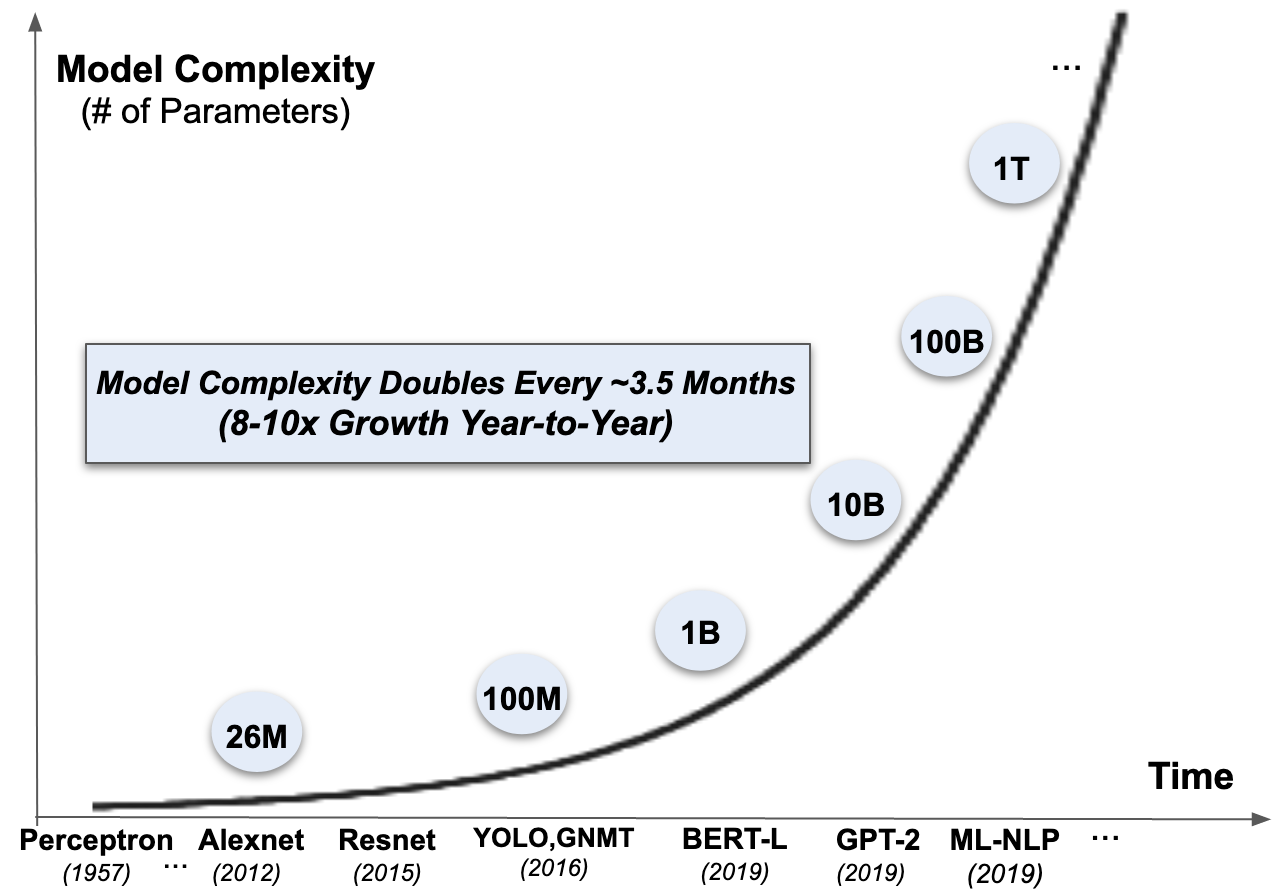}
\setlength{\belowcaptionskip}{-8pt} 
\caption{\small{Model complexity continues to grow at an exponential rate \citep{AISummit_Intel_Keynote}.}}
\label{fig:model-complexity}
\end{figure}
AI model complexity is increasing at an unprecedented rate \citep{ChengWZZ2017}.
Metrics for the number of building blocks as well as the complexity within each of the building blocks point to consistent increases.
For example, the number of AI model parameters increases almost tenfold annually \citep{AISummit_Intel_Keynote},
which is much more aggressive than the traditional Moore's Law scaling.
As the number of parameters approach one trillion, such complex models motivate a reevaluation of fine-grained or parameter-level governance reviews.

\subsection{Agility and Time to Review}
\smallskip

\begin{challenge}
AI governance reviews can often take a year or longer.
\end{challenge}
Most model governance guidelines provide optimistic timelines for AI governance reviews (on the order of weeks or months depending on the firm).
However, in practice, the end-to-end AI governance review process frequently takes close to a year or longer.
The length of the process is driven by factors like  complex review flows and the lack of automation.
The number of models in large financial firms is increasing by {10--25\%} according to recent reports \cite{mckinsey_modelcount}.
The length of the review has implications on the successful adoption of AI, cost, operational efficiency, and model performance.
Recent months have further highlighted the importance of time-to-review and time-to-deploy metrics for robust and effective modeling processes.  

\begin{challenge}
Long governance reviews negatively impact the AI model performance and agility.
\end{challenge}
One unintended consequence of long model governance reviews is degraded model performance.
AI models are typically locked during the governance reviews and start aging, along with underlying data and assumptions.
Model aging becomes a major concern as review times lengthen.
The effects are even more prominent in dynamic or adversarial environments,
e.g., in fraud detection,
where model agility is crucial for effectiveness. 

\subsection{Regulatory Complexity and Uncertainty}

\begin{challenge}
Regulatory complexity and uncertainty make governance increasingly difficult.
\end{challenge}
As discussed earlier, financial institutions are expected to comply with numerous regulations.
For US credit models, these may include the Fair Housing Act, Consumer Credit Protection Act, Fair Credit Reporting Act, Equal Credit Opportunity Act, Fair and Accurate Credit Transactions Act, as well as regulatory directives such as Regulation B.
Furthermore, these regulations are enforced by different regulators at the international, federal, state, and local levels, as summarized in \Cref{fig:regulators} \citep{R44918}.
In recent years, AI and data regulations have become more rigorous and added to the underlying financial regulatory complexity (such as GDPR and CCPA).
While regulatory systems and complexity serve numerous purposes, building compliant AI models in this complex regulatory landscape is a formidable task.
Up to 67\% of the financial institutions in the US and EU perceive regulatory complexity and uncertainty as concerns \cite{wef2020}.
More importantly, up to 94\% and 97\% of financial firms in AI leadership positions indicate seeing regulatory complexity and uncertainty respectively as problems, which is remarkably higher than the perceptions by the laggards in terms of AI maturity. Increased automation and system-level support are essential in addressing these concerns.

\begin{figure}
\centering
\includegraphics[keepaspectratio,width=0.75\columnwidth]{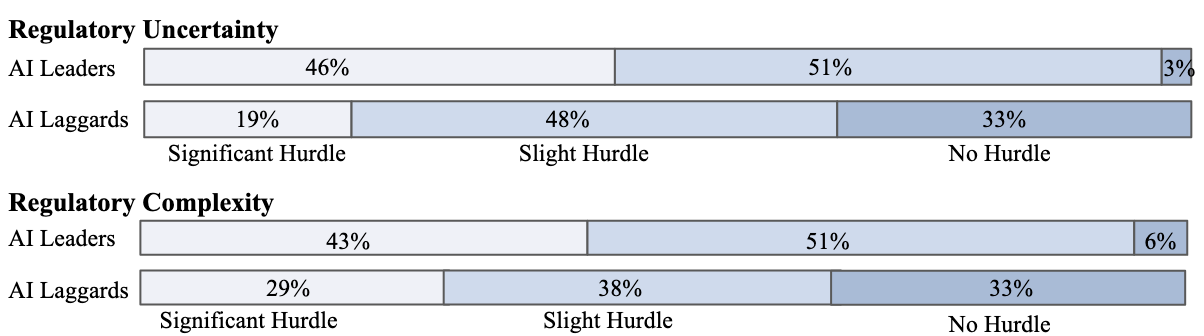}
 \setlength{\belowcaptionskip}{-4pt} 
\caption{\small{The perception of financial regulatory complexity and uncertainty by AI program maturity \textit{\small{(source statistics from\citep{wef2020})}}}.}
\label{RegulatoryComplexity}
\end{figure}

\begin{challenge}
Regulatory complexity incentivizes AI models to comply with the most restrictive regulatory guidelines.
\end{challenge}
Regulatory complexity has multiple dimensions, including the types of applications, regulation types, regulatory entities, jurisdictions such as local, state, country, and territory levels, each with their own unique requirements
In response to the resulting complexity, it is easiest to build models that adhere to the largest possible set of regulations, especially as new regulations in one jurisidiction inspire adoption elsewhere \citep{Chin2019}.
For instance, if an AI model is deployed in multiple territories, where one territory has more conservative guidelines, model governance would prefer a model that is designed for the most conservative guidelines across all territories.
Since the current processes do not have any separation of regulatory functions from the AI models, model performance typically gets impacted by these additional restrictions. 

\subsection{Modeling Restrictions}

\begin{challenge}
Governance restricts the AI models to conservative modeling approaches.
\end{challenge}
In an attempt to reduce the AI risks, governance practices commonly constrain the models to conservative modeling approaches. Restrictions on the modeling framework include offline-only training, model types (such as restrictions on the use of neural networks), etc.
Despite showing promising results in many application areas, many modeling techniques have largely not been allowed in financial services under current governance guidelines.
Such strict restrictions do not always produce lower risk outcomes, e.g., when the application characteristics mismatch with modeling assumptions. Even limited and conditional use of emerging approaches may provide strategic, long-term benefits, e.g., reinforcement, online, incremental, and active learning \citep{MinhKSR2015,LoboSBK2018,ChalapathyMC2018}.

\begin{challenge}
AI governance usually assumes static or near-static environments, which may be unrealistic.
\end{challenge}
AI model governance processes frequently enforce the use of static or near-static environment assumptions for the approved model use period \cite{BPI2020}. Even though AI models do not rely on assumptions, in some cases the static assumptions are established by restricting the model training or use periods to approximate the static conditions. The mismatch in the assumed and real conditions causes performance problems. As an example, in payment systems, fraud scoring models are expected to assume near-static transactions and fraud patterns.  In reality, fraud patterns often go through major changes in short periods of time. In person-to-person (P2P) payments the fraud tactics can change in a matter of days and cause significant losses as the underlying models are not equipped to deal with dynamic environments. Furthermore, static assumptions prevent models from dealing with unexpected changes and limit robustness. 

\subsection{Regulatory Monitoring and Reporting}

\begin{challenge}
Intermittent monitoring fails to identify critical changes in the environment, data drifts, or data quality issues.
\end{challenge}
Financial institutions are required to perform on-going monitoring reviews at least on an annual basis. For many AI models, the on-going monitoring frequencies remain close to these baseline requirements.  The lack of continuous monitoring causes the critical environment changes, data drifts, and data quality issues remain undetected. This, in turn, has performance, robustness, and compliance implications. 

\begin{challenge}
Traditionally, model monitoring has been performed on a limited number of metrics.
\end{challenge}
Model development and governance processes have mostly been built with static environment assumptions. In recent months, due to effects of the global pandemic, the requirement to demonstrate mostly static operational environments for AI models has been reevaluated by many financial services firms.  There has been significant interest in monitoring a broader range of metrics, to capture the changes in the underlying data, model behavior and to ensure robustness, even for adaptive models.


\begin{challenge}
Regulatory metrics are growing faster than the capacity of the manual governance processes.
\end{challenge}
A growing number of metrics are being proposed and used for regulatory purposes \cite{BLDS2020}, such as quantifying bias in credit applications using demographic parity \citep{Chouldechova2016}, equalized odds \citep{HardtPS2016,Corbett-DaviesPFG2017,GolzKP2019},
and other group fairnesses \citep{kim2020modelagnostic}, Adding more and increasingly sophisticated metrics to the monitoring list is not scalable, due to manual data extraction, reporting, committee reviews.
The metrics themselves may even be mutually contradictory \citep{PleissRWK2017,kim2020modelagnostic} or be difficult to operationalize.

\begin{challenge}
Reporting performance and monitoring results to the appropriate committees is a manual and laborious process.
\end{challenge}

\noindent During model deployment, the development team is responsible for collecting the data or metrics of interest, producing a written report using predetermined templates, and presenting it to numerous committees.
Stakeholders or internal committees do not have access to the deployment system and its behavior other than this indirect, committee review path.
Depending on the model, this manual process may take a few weeks to months.
The inherent complexity and length of the reporting process limit the frequency and effectiveness of monitoring.

\subsection{Mitigation and Regulatory Control}

\begin{challenge}
Regulatory issues are not detected or rectified promptly due to the lack of run-time monitoring or mitigation.
\end{challenge}
When monitoring is insufficient, regulatory breaches may be discovered only during the model retraining exercises or through regulatory inquiries, by which time an institution is at risk of incurring fines, penalties and reputational damage \citep{Armour2017}.
This highlights the need to perform continuous monitoring as well as run-time remediation.
Using a robotics analogy, the current practices would translate to not including any real-time monitoring (camera) or control (steering or other actuation feedback) in the broader system. 


\subsection{Robustness and Stress Testing}

\begin{challenge}
Despite being commonly used for traditional financial models, stress tests are not available for broader AI use cases.
\end{challenge}

\noindent Traditional models such as capital analysis rely heavily on the industry-wide standard stress tests.
At this point in time, no such tests are available for most AI models in financial services.
Model development organizations often try to mimic aspects of the stress tests by experimenting with limited and theoretical scenarios.
However, as discussed in \Cref{sec:intro}, the nature of AI models require extensive scenario and stress testing for robustness and compliance purposes.

\begin{challenge}
Model development organizations face serious problems in reassuring governance committees about compliance.
\end{challenge}
Model development teams face genuine difficulties in assuring risk management organizations that the AI models are on target for compliance, robustness, and risk criteria.
This translates to an open-ended burden of proof for the modeling teams.
It involves justifying all model parameters, modeling decision, data, features, assumptions, architecture, and techniques. Yet, even after many rounds of reviews and additional scrutiny, AI model compliance still remains unclear.

\section{System-level Framework Towards increased AI Self-Regulation}
\label{sec:controller}

The long list of AI governance challenges motivates the need for new solution approaches.  In this section, we envision and present a high-level AI system framework and modular building blocks towards increased self-regulation and more efficient AI governance in financial services.
The proposed framework lies at the boundary of the AI and governance systems. It is not intended as a specific design blueprint, but a high-level design approach that can be customized based on the AI model, application, and the firms' needs.



\begin{figure}
\centering
\includegraphics[keepaspectratio,width=0.95\columnwidth]{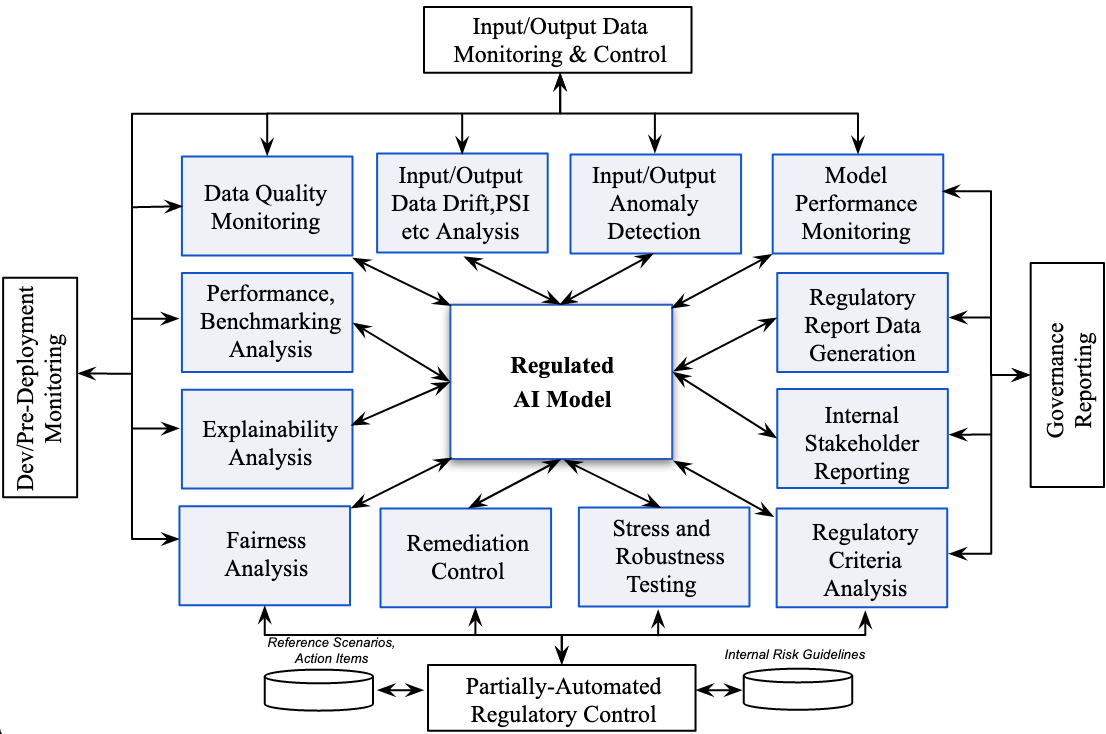}
 \setlength{\belowcaptionskip}{-6pt} 
\caption{\small{High-level view of the AI system framework with self-regulatory capabilities.}}
\label{fig:self-govern}
\end{figure}

\begin{capability}
Continuous regulatory monitoring and reporting during deployment
\end{capability}

\noindent \Cref{fig:self-govern} shows the high-level architecture of the proposed AI system framework and regulatory modules. The system aims to incorporate run-time monitoring, regulatory control, and mitigation capabilities in the production environment. The figure shows a broad selection of modules for completeness, yet a smaller subset of modules can be customized for each application, such as mortgage applications or credit card payments.
Monitoring capabilities are intended for self-regulation purposes. The outputs of the monitoring system ties to regulatory functions to ensure that the system behaves within guidelines. 

Monitoring functions may include: (i)\textit{model vital statistics:} essential metrics and statistics of the model's behavior and data (such as volumes, data quality metrics, etc.); (ii)\textit{input data:} monitoring of the input streams (such as fairness metrics, data drift, population stability metrics etc.);  (iii) \textit{extended data monitoring:} analysis of the model's input/output behavior using extended data sources, like features that the AI model is not allowed to access, correlation of the input features with protected classes such as age, race, gender, etc.; (iv) \textit{output monitoring:} monitoring for performance, fairness, and other regulatory criteria; (v) \textit{emerging pattern and anomaly detection:} continuous monitoring of the abnormal and emerging patterns in the input/output streams, as well as extended data streams; (vi) \textit{automated report generation:} reporting for development teams, governance committees, or regulators, such as regulatory metrics, alert frequencies, alerting levels, compliance confidence scores, etc.

\begin{capability}
Integration of key self-regulatory building blocks
\end{capability}

\noindent The framework enables the integration of modular, self-regulatory blocks in the AI system itself to cover common requirements, without increasing the complexity of the core AI models. These blocks are expected to be: (i) customized for the application types, (ii) used by multiple application and model types with customizations.  As discussed earlier, the specific list of modules is bespoke for each application type based on regulatory guidance. Modular blocks provide a natural way to separate the regulatory criteria from the AI model, while keeping them in the same system. This enables the system to integrate and update governance blocks independent of the AI model. It ensures improved compliance for a wider range of models without having to rebuild the regulatory functionalities within each model instance.  AI specific analysis modules can be incorporated in the system: (i) \textit{explainability analysis:} for regulatory and customer inquiries as well as audit purposes; (ii) \textit{data quality analysis:} to continuously monitor data quality issues (such as missing, unexpected, out-of-range data, systemic issues and failures etc.); (iii) \textit{fairness analysis:} monitoring and mitigating bias (such as fairness metrics, input correlations with protected classes, etc.
Recent policies and regulatory publications emphasize and clarify the transparency and explainability requirements in adverse action notices \cite{CFPB}.

The lack of universal regulatory guidelines remains a notable challenge, and will likely be overcome through regulatory leadership.
However, each financial institution has its own internal guidelines based on the combination of the firm's risk appetite and regulatory guidelines. In this study, we propose using internal guidelines for the framework in the near term.
As an example, a commonly used metric in bias is the prediction accuracy delta for different classes, for which the specific thresholds can be customized by each firm based on their risk appetite. Once implemented, the proposed system also has the opportunity to enforce these internal regulatory metrics pervasively across many AI models. 

\begin{capability}
Reusable libraries of module templates, system and regulatory guidelines
\end{capability}
\noindent The development of the proposed governance modules involves: (i) Libraries of module templates for different application and regulation types, such as bias/fairness, explainability, data quality templates with baseline functionalities. In some cases the module libraries may include full models as building blocks; (ii) Customization and configuration of the module templates by the development teams, such as customizing the metrics, specifying the input features, outputs, etc.; (iii) Criteria selection based on the firms risk appetite, application type and regulations, such as the conditions and quantitative metrics for governance criteria, alerting and mitigation; (iv) Guidance on system architecture requirements, such as the required list of modules, metrics and functionalities for an application type; (v) Libraries for control scenarios and mitigation actions are expected to be gradually built for application types by the model governance and development organizations.

\begin{capability}
Run-time mitigation with human-in-the-loop alerting
\end{capability}
\noindent The system-level framework aims to gradually build automated mitigation capabilities. In early stages, it targets partially automated run-time mitigation and human-in-the-loop alerting. The goal is to manage the model's behavior more effectively during run-time. The system does not aim to provide compliance guarantees. The approach provides a number of orthogonal capabilities: (i) Scenario-based mitigation for well-defined control paths, for which the scenarios can be extracted through the use of continuous monitoring data from pre-deployment testing, historical data, and established scenarios such as abnormal payment transaction patterns during holidays;  (ii) System-level remediation through the use of alternative models, such as using shadow AI models outputs for population segments if the primary AI model shows detectable bias during continuous monitoring; (iii) Post-processing and re-calibration of the model outputs, such as calibration of model scores to compensate for detected issues.

\begin{capability}
Governance controller and AI-based governance
\end{capability}
For increased automation in some cases, a \textit{governance controller} may be used, along with human-in-the-loop validations of control actions. The controller interacts with the individual modules through module output monitoring, alerting, and mitigation actions. It may select the output from a collection of alternative AI models to ensure the system-level compliance and robustness. Furthermore, the controller enables the run-time management of the model according to the firms' risk policy guidelines.
In the later stages of adoption, AI itself will play a big role in the governance and regulatory functions. This involves 2 orthogonal paths: (i) intrinsically self-regulating AI models; (ii) AI-based external governance and regulatory control. The proposed approach is essential for both paths, providing key capabilities in data and monitoring, risk and regulatory guidance, required metrics and criteria etc. 

\begin{capability}
Dynamic reconfiguration for change management, model replacement and scale-out
\end{capability}

\noindent The inherent modularity of the system-framework enables plugging in alternative regulatory modules, reconfiguring or changing them independent of the core AI model. As regulations are updated, corresponding modules can be updated across the firm without retraining the individual AI models. This is essential to ensure compliance during the lifetime of the AI system when: (i) a new AI model is developed to replace an older model; (ii) the AI model is to be scaled out and used in multiple territories with different regulations or regulatory variations; (iii) the regulatory guidelines change over time as new regulations are introduced.

\begin{capability}
Governance throughout the model lifecycle 
\end{capability}
One of the unique capabilities that the system-level approach aims to offer is to provide a regulatory framework throughout the entire model lifecycle, where the capabilities may be customized for individual application types. As they are template based, governance modules can be utilized in different platforms during pre-deployment testing or development.  The pre-deployment modules may closely mimic deployment versions, where development functionalities and modules may include: (i) Data quality modules and guidelines; (ii) Regulatory criteria, such as bias and explainability; (iii) Performance metrics, guidelines and benchmarking requirements for individual applications. Through the increased automation of development and pre-deployment governance functions, initial model approval process can be improved significantly.  

\begin{capability}
Custom robustness tests for the AI model
\end{capability}
Continuous monitoring provides the opportunity to collect vast amounts of run-time behavioral data. The proposed system uses this resource to develop custom robustness and stress tests for the AI model. This relies on identifying the weaknesses, failure patterns, and risky scenarios/patterns during continuous monitoring.  These model-specific scenarios can be extended with general scenarios,  such as those identified in earlier or alternative models.
The outcomes of these tests directly tie into the mitigation functions in the proposed system-level framework.
In addition, the framework provides the ability to integrate externally developed stress-tests from regulators or vendors, as they become available.
Robustness tests and stress tests are valuable tools in assuring the model risk management teams about compliance. These capabilities have the ability to refocus the governance review processes from the parameter-level justifications to the system-level targets with quantifiable outcomes.
It also helps assure the AI governance teams by showing a reasonable number and range scenarios are covered through these analyses.
In later stages, the generative adversarial modeling capabilities may be used for advanced and continuous stress tests.

\section{Solution Opportunities}
\label{sec:opportunities}
The system-level approach provides novel capabilities and solution opportunities:

\begin{Opportunity}
Develop customized and scalable solutions for robust AI systems and governance challenges.
\end{Opportunity}
The system-level approach provides the opportunity to alleviate majority of the pain points discussed in \Cref{sec:challenges} by integrating key capabilities in the AI system itself. This allows improvements in performance, cost, complexity, time-to-develop, time-to-review, and the ability to scale out. It enables the robust operation of the models as well as providing a new \textit{run-time risk mitigation} path for model risk management and governance organizations. The combination of increased model and regulatory complexity raises questions on the sustainability of the current practices. The solution approach provides a sustainable solution under the dual forces of complexity.

\begin{Opportunity}
Integrate key governance capabilities in a unified, in-house framework with increased automation.
\end{Opportunity}
Governance and compliance solutions frequently suffer from fragmented environments, inconsistent tools, and techniques. The self-regulating AI system approach provides a unified in-house governance framework as well as the ability to customize for different application types. It enables incorporating the third-party solutions in the same architecture with the in-house solutions, while universally managing all based on the firms' risk management policies and guidelines. 

\begin{Opportunity}
Enable the run-time management and mitigation of the models.
\end{Opportunity}
Financial institutions have largely avoided the run-time management of AI models. The system-level approach aims to provide real-time monitoring and mitigation capabilities to effectively manage AI models in production. The monitoring and mitigation capabilities can be built up from the basic human-in-the-loop alerting to partially or fully automated control and controller architectures. AI models will likely play a big role in the regulatory and governance functions in near future. The proposed framework provides enabling capabilities and data sources for AI-based governance solutions.

\begin{Opportunity}
Implement firms' risk policies and regulatory guidelines pervasively and at scale.
\end{Opportunity}
The libraries of governance module templates, architectural requirements and regulatory criteria provide the opportunity to uniformly implement the firms risk management policies and the regulations at scale. The guidelines on the regulatory module selection for each application, the contents of the module templates, thresholds and criteria can be pervasively implemented for all AI models.

\begin{Opportunity}
Enable the development of custom robustness tests.
\end{Opportunity}
The lack of stress tests and run-time monitoring are serious challenges in managing the risks of AI solutions. Continuous run-time monitoring provides the opportunity to develop robustness tests customized for the model. This shifts the AI model governance focus from the existing parameter-level processes to an empirical and system-level approach.

\begin{Opportunity}
Enhance risk management throughout the model lifecycle.
\end{Opportunity}
AI governance reviews mostly focus on the pre-deployment and deployment through the initial model approval and on-going monitoring processes respectively. With the availability of the libraries of governance modules and tools, the corresponding functional capabilities can be integrated in earlier stages of the model development cycle and in different platforms, which can significantly improve the end-to-end process. Finally, the proposed approach provides a new risk management stage during the deployment of the model, through continuous monitoring and mitigation. 

\section{Adoption Considerations}
\label{sec:Adoption}
This section discusses some of the key adoption considerations for the proposed framework.

\textit{Managing Expectations:}
Autonomy, trustworthiness and robustness remain among the list of AI grand challenges. The proposed system does not offer a panacea, but rather a step in improving the overall governance process and enabling solution opportunities.

\textit{Organization and Process Enhancements:}
Without any internal process adjustments, it is not possible to achieve the full potential of the governance framework. Possible process improvement opportunities include: reducing the number and complexity of the reviews, simplifying the roles and responsibilities of the numerous committees, providing insights to the committees via direct access to monitoring and compliance metrics.  

\textit{Development of Libraries and Module Templates:}
The proposed framework relies on libraries and guidelines to be provided by the governance organizations.
Similarly, the customization and integration stages are expected to be performed by development teams.
Addressing these needs requires significant initial investment in the system, which should yield long-term payoff. Gradual integration of these capabilities is a reasonable approach to achieve the end-state of the system. 

\textit{Regulatory Standards and Guidelines:}
Recent studies agree on the need for standard regulatory guidelines for AI \cite{GoogleAI,HarvardAI,HKIMR}. In this study, we acknowledge the challenges in achieving such universal guidelines and rely on financial firms' internal risk guidelines as an alternative for the time being.  While some of the internal standards are already available, others may be added gradually over time. 
 
\textit{Platform and Deployment Considerations:}
The proposed system-level approach is mostly platform-agnostic. While the primary application of the proposed system is the deployment platforms, the modules and capabilities can be utilized both in pre-deployment and deployment platforms, while the underlying templates can be extended to development platforms. 

\section{Conclusions and Outlook}
\label{sec:conclusion}
As AI systems have become integral parts of the financial services industry, it is essential to have effective model governance practices to ensure robustness and compliance. Legacy governance processes in practice today suffer from performance, cost, complexity, agility, and scaling issues in AI models. This paper tries to focus on some of the common challenges in AI model governance in the financial services industry. Given the unprecedented growth in AI complexity, the feasibility of the current governance practices for the next generation of AI models is questionable. The paper aims to start an initial conversation on the key challenges and solution opportunities. A system-level framework with modular building blocks is presented.   This approach aims to incorporate increased automation, integration and configurability into the AI system towards self-regulation. It enables some key capabilities to alleviate the existing challenges and enable more effective and compliant AI solutions. 

{\smaller
\begin{spacing}{0.9}
\paragraph{Disclaimer}
This paper was prepared for informational purposes in part by the Artificial Intelligence Research group of JPMorgan Chase \& Co and its affiliates (``JP Morgan''), and is not a product of the Research Department of JP Morgan.  JP Morgan makes no representation and warranty whatsoever and disclaims all liability, for the completeness, accuracy or reliability of the information contained herein.  This document is not intended as investment research or investment advice, or a recommendation, offer or solicitation for the purchase or sale of any security, financial instrument, financial product or service, or to be used in any way for evaluating the merits of participating in any transaction, and shall not constitute a solicitation under any jurisdiction or to any person, if such solicitation under such jurisdiction or to such person would be unlawful.
\end{spacing}
}

\bibliography{bib}
\bibliographystyle{ACM-Reference-Format}
\end{document}